%% file: main.tex
\documentclass[runningheads]{llncs}
\usepackage{amsmath}
\usepackage{graphicx}
\usepackage{subcaption}
\usepackage{amsfonts}
\usepackage{multirow}
\usepackage{color}
\usepackage[dvipsnames,table,xcdraw]{xcolor}
\usepackage{xspace}
\usepackage{booktabs}
\usepackage{algorithm}
\usepackage{algorithmic}
\usepackage[hidelinks]{hyperref}

\input{preamble}

\begin{document}
\title{Federated Document Visual Question Answering: A Pilot Study}
%
%
\author{Khanh Nguyen\inst{1}\orcidID{0000-0003-3353-0473} \and
Dimosthenis Karatzas\inst{1}\orcidID{0000-0001-8762-4454} }

\authorrunning{K. Nguyen et al.}
%
\institute{Computer Vision Center, Universitat Autonoma de Barcelona, Spain
\email{vkhanh,dimos@cvc.uab.es}}
\maketitle              
\begin{abstract}
An important handicap of document analysis research is that documents tend to be copyrighted or contain private information, which prohibits their open publication and the creation of centralised, large-scale document datasets. Instead, documents are scattered in private data silos, making extensive training over heterogeneous data a tedious task.
In this work, we explore the use of a federated learning (FL) scheme as a way to train a shared model on decentralised private document data. We focus on the problem of Document VQA, a task particularly suited to this approach, as the type of reasoning capabilities required from the model can be quite different in diverse domains. Enabling training over heterogeneous document datasets can thus substantially enrich DocVQA models.
We assemble existing DocVQA datasets from diverse domains to reflect the data heterogeneity in real-world applications. We explore the self-pretraining technique in this multi-modal setting, where the same data is used for both pretraining and finetuning, making it relevant for privacy preservation. We further propose combining self-pretraining with a Federated DocVQA training method using centralized adaptive optimization that outperforms the FedAvg baseline. With extensive experiments\footnote{The code is available at ~\url{https://github.com/khanhnguyen21006/fldocvqa}}, we also present a multi-faceted analysis on training DocVQA models with FL, which provides insights for future research on this task. We show that our pretraining strategies can effectively learn and scale up under federated training with diverse DocVQA datasets and tuning hyperparameters is essential for practical document tasks under federation.

\end{abstract}

\section{Introduction}
Most contemporary documents relevant for document analysis research tend to contain private information or be under copyright restrictions. This includes everything from administrative documents, bank statements, insurance claims, identification documents, student exams, bills, invoices, books, magazines, etc. Moreover, several data regulation policies (e.g. the General Data Protection Regulation (GDPR) in Europe or state consumer privacy acts in the US) further restrict the way such documents can be used and distributed.

As a result, in various real-life scenarios data is collected, stored and processed locally, and cannot be pulled together and used for training large-scale models. Subsequently, most research is performed either on outdated document collections, or on small-scale, private datasets, which rarely reflect the heterogeneity of the real-world, and render the research outcomes difficult to reproduce.

In this work, we explore for the first time the use of Federated Learning for training a large multimodal LLM model for Document Visual Question Answering (DocVQA). Figure \ref{fig:fedocvqa} shows an overview of our approach.
\input{figures/teaser}
Federated Learning \cite{mcmahan2017communication} is an alternative training paradigm, that enables a set of distributed clients to collaboratively train a centralized model without sharing their local data. It has been used extensively in fields such as health \cite{dayan2021federated}, recommendation systems \cite{lin2020fedrec}, smart cities \cite{jiang2020federated}, etc, but to the best of our knowledge it has not been applied yet to document understanding.

The DocVQA task gave rise to a new prompt-response based scheme for document understanding. In this scheme, the input is invariably multimodal, comprising on one hand the question and on the other the visual representation of the document. In most cases, the document visual representation is coupled with the OCR and layout data, making the input space even more complex.

The majority of previous Federated Learning (FL) work has focused on the unimodal setting, where the local data of all clients is described by a single modality. In contrast, DocVQA is a case of congruent multimodal FL \cite{che2023multimodal}, where the clients hold multimodal data with the same local modality combination, in our case: vision and language. This introduces further challenges to the FL setup.

In the DocVQA scenario, the FL setup needs to resolve the heterogeneity challenge in terms of data distribution (i.e. the fact that data distribution is not balanced between the clients), while at the same time efficiently leverage domain heterogeneity across the clients. In the case of DocVQA, the basic question-answering scheme is the same, but each client has access to documents (and corresponding questions) from different domains.

Notably, similarly to the fact that training (fine-tuning) data is found in data silos, representative pre-training data is also difficult to get hold of in a centralized location. To solve this problem, we use self-pretraining in a federated learning setting. Self-pretraining uses the same data both for pre-training (without annotations) and for the final training step (with annotations).

In this work we study the application of Federated Learning to DocVQA, a complex DU problem with multimodal input. To the best of our knowledge, this is the first study of FL application to the document analysis and recognition field. Specifically, our contributions are the following:
\begin{enumerate}
    \item We provide a new benchmark for Federated Learning DocVQA, with a well-defined non-IID distribution, bringing together 3 existing DocVQA datasets to simulate a realistic setting for FL.
    \item We introduce a new training setup that employs FL for DocVQA, where we train a model on private, distributed documents without centralized data aggregation, to collaboratively train a model on multiple DocVQA datasets from diverse domains.
    \item We explore the effectiveness of various FL optimization strategies at the server side, and report results on different methods.
    \item We propose the use of self-pretraining in our proposed Federated Learning setup and demonstrate its effectiveness.
    \item We show that the use of server adaptive optimization, combined with self-pretraining improves DocVQA performance compared to the FedAvg baseline approach, reaching results to the centralized training method.

\end{enumerate}
\section{Related Work}
\subsection{Document Visual Question Answering}
DocVQA has gained significant attention in the field of DU due to its nature of answering natural language questions to extract information from documents. As a result, numerous datasets addressing diverse domains and challenges have emerged, including industry documents~\cite{mathew2021docvqa,tito2021icdar}, infographics~\cite{mathew2022infographicvqa}, structured tables and databases~\cite{pasupat-liang-2015-compositional,Chen2020TabFact:}, multipage~\cite{tito2023hierarchical}, multidomain~\cite{van2023document}. 
Despite the availability of small and medium-sized datasets, the lack of a large-scale generic dataset suitable for diverse scenarios persists. This challenge is primarily due to the sensitive content and copyright issues surrounding many documents, which prevent document holders from publicly releasing their data. In this context, Federated learning techniques, serving as privacy-preserving methods, offer a solution to facilitate the use of distributed and private datasets among different entities.

Recent advancements in single-page DocVQA, such as LayoutLM family \cite{huang2022layoutlmv3}, TILT \cite{powalski2021going}, UDOP \cite{tang2023unifying} to name a few, have achieved impressive results through the integration of OCR, layout and visual features. Another research works like Donut \cite{Kim22Donut} and Pix2Struct \cite{Lee23Pix2Struct} introduce end-to-end architectures that release the burden on external OCR modules by incorporating reading-oriented pre-training objectives, maintaining competitive performances. 
The key component of these methods is the use of self-supervised pretraining objectives on  large-scale document datasets, resulting in better representations for downstream tasks.
In this work, we base our architecture on a text-only pre-trained language model (PLM) T5 \cite{raffel2020exploring} as the backbone, enhanced with visual features extracted from the document to accommodate multimodal input. This method offers ease of fine-tuning along with robust performance on the DocVQA task \cite{borchmann2021due}, allowing us to focus on privacy-preserving techniques like FL on multimodal data.

\subsection{Cross-silo Federated Learning}
The cross-silo Federated Learning \cite{kairouz2021advances} is relevant when multiple companies or organizations have a shared interest in training a model using their collective data, but are unable to share their data directly due to confidentiality, legal or geological constraints. In this setting, data can be partitioned either by examples (and)or features, denoted as \textit{horizontal}/\textit{vertical} FL \cite{yang2019federated}, respectively. In horizontal FL, training data from various parties have different distributions but share the same feature space. In contrast, vertical FL refers to settings where data from different parties share the same distribution but have differing features.

The horizontal setting where data is example-partitioned is our main focus in this work. Currently, proposed FL algorithms mostly focus on addressing challenges from non-IID and unbalanced data setting, by synthesizing client-partitioned datasets \cite{li2021model,hsu2019measuring} from single modal tasks, resulting in better optimization algorithms \cite{mcmahan2017communication,li2020federated,karimireddy2020scaffold,reddi2020adaptive}. Concurrently, another research lines in FL tackle problems in communication efficiency \cite{hamer2020fedboost,sattler2019robust}, compression \cite{tang2019doublesqueeze,vogels2019powersgd} and differential privacy with FL \cite{pmlr-v130-girgis21a,ghazi2019scalable}. We take a step further to extend this setting to a more realistic multimodal document system.
\subsection{Self-Pretraining}
\textit{Self-Pretraining} is an accepted notion in NLP where the same (downstream) training data is utilized for both pretraining and finetuning \cite{krishna2022downstream}. In many applications, best practices involve pretraining models on large unlabeled \textit{upstream} datasets using self-supervised objectives, followed by finetuning on labeled \textit{downstream} data for the specific task of interest. These large-scale pretrained models often yield significant improvements compared to models trained solely on the downstream task.

However, recent studies in NLP \cite{gururangan2020don,krishna2022downstream,wu2022insights} exploring the impact of large-scale corpora have revealed surprising results: notable performance gains are observed even when using synthesized upstream data \cite{wu2022insights} or identical upstream and downstream datasets \cite{krishna2022downstream}, as long as the pretraining objective captures task-related structures. This phenomenon also extends vision models, where pretraining with a Masked AutoEncoder (MAE) \cite{he2022masked} objective has been shown to enhance the performance of ViT \cite{dosovitskiy2020image} on ImageNet-1K \cite{5206848}, and pretraining solely on downstream datasets for object detection and segmentation has achieved performance comparable to ImageNet-pretrained models \cite{el2021large}.
Despite of these interesting findings, several practical questions related to document tasks remain unexplored: (1) the extent to which self-pretraining is advantageous with limited-scale downstream data and high-level reasoning pretraining objectives, and (2) whether this observation is domain/task-agnostic and can be generalized to other complex domains. 
In this study, we re-visit this idea in a multimodal setting like DocVQA and show that continuing pretraining PLMs on unlabeled documents in downstream DocVQA datasets improves the DocVQA performance of finetuning in the same tasks.
\section{Methodology}
\subsection{Problem Statement}
This section provides the formal definition of Federated DocVQA, termed FeDocVQA. Suppose there are $K$ relatively reliable clients. Each client $k$ holds a DocVQA dataset $D_k$ of moderate size $n_k$ and guarantees a sufficient computational capability. The goal is to collaboratively train a DocVQA model $\theta$ among clients \textit{without data exchange}. The training process, which is coordinated by a central server, minimizes the objective function:
\begin{equation}
\begin{aligned}
    \min_{\theta \in \mathbb{R}^{d}} f(\theta) := \sum^{K}_{k=1}{p_{k}F_{k}(\theta)}\\
\end{aligned}
\end{equation}
where $F_{k}(\theta) = \mathbb{E}_{x \in D_{k}}[\mathcal{L}_{k}(\theta;x)]$ is the typical DocVQA training loss with training examples $x\sim D_k$, as defined in \cite{mathew2021docvqa}. In real-world FL scenario, for any $i\neq j$, $D_i$ and $D_j$ are usually very different. We also have $p_{k}=\frac{n_k}{\sum^{K}_{j=1}n_j}$ is the assigned weight to client $k$ that controls its impact during training, which is proportional to the local data size $n_k$. 
\subsection{Federated Learning}
\label{sec:fl}
We now present a general recipe for federated training algorithms called FedOPT \cite{reddi2020adaptive}, that includes FedAvg \cite{mcmahan2017communication} and many others \cite{li2020federated,hsu2019measuring,karimireddy2020scaffold} in the literature, as summarized in Algorithm \ref{alg:fedopt}. Particularly, a server coordinates the training procedure by repeating the following steps in $T$ communication rounds. The training starts with a global model initialized as $\theta_0$. At round $t$: 
\begin{enumerate}
    \item The server randomly samples a set of clients $\mathcal{S}^{t}$ with a probability $C$ without replacement, assuming that clients are always available. As a result, $C\cdot K$ clients are selected to train the model with its DocVQA data in each round.
    \item  Server sends the current model weights $\theta_{t}$ to the selected clients. All clients agree to follow the same training algorithm and objective, defined as CLIENTOPT. It is also assumed that all clients are able to perform the entire local training that is required. 
    \item Each sampled client trains the model with its local data $D_{k}$ for $E$ epochs via CLIENTOPT. Then its update, which is the difference in model's parameters before and after local training, is sent back to the server.
    \item The server collects an aggregation of updates $\Delta^{t}$ from clients participated in the current round. 
    \item The server updates the global model following a gradient-based optimization strategy, defined as SERVEROPT, where the aggregated update $\Delta^{t}$ serves as pseudo server gradient.
\end{enumerate}
\begin{algorithm}[tb]
   \caption{\textbf{FedOPT}}
   \label{alg:fedopt}
    \begin{algorithmic}
       \STATE {\bfseries Input:} Initialize global model $\theta_{0}$, server/client learning rate $\eta_{s}$/$\eta_{l}$, local epochs $E$, rounds $T$
       \FOR{each round $t=0$ {\bfseries to} $T-1$}
        \STATE Sample a subset $\mathcal{S}^{t}$ of clients \quad // \textit{Step 1.}
        \STATE Server sends $\theta_{t}$ to all clients in $\mathcal{S}^{t}$  \quad // \textit{Step 2.}
        \FOR{each client $k \in \mathcal{S}^{t}$ {\bfseries in parallel}} 
        \STATE Initialize local model $y^{0}_{k}\leftarrow \theta_{t}$, 
        \STATE Client $k$ performs $E$ epochs of local training via $y^{j+1}_{k} = \text{CLIENTOPT}(y^{j}_{k}, \eta_{l}, t)$, $j$ denotes the gradient step.
        \STATE Let $y^E_{k}$ denote the local model after $E$ epochs of local updates
        \STATE Client $k$ sends $\Delta^{t}_{k} = \theta_{t} - y^E_{k}$ to Server
        \ENDFOR  \quad // \textit{Step 3.}
       \STATE Server aggregates client updates $\Delta^{t} = \frac{1}{|\mathcal{S}^{t}|}\sum_{k\in\mathcal{S}^{t}}p_{k}\Delta^{t}_{k}$  \quad // \textit{Step 4}.
       \STATE $\theta_{t+1}\leftarrow \text{SERVEROPT}(\theta_t, \Delta^{t}, \eta_{s}, t)$  \quad // \textit{Step 5.}
       \ENDFOR
    \end{algorithmic}
\end{algorithm}
In our approach, we fix the training algorithm \text{CLIENTOPT} at all clients while examining the effectiveness of various optimization strategies for \text{SERVEROPT} at server side, as described below. Given the pervasive use of PLMs in DocVQA (details in Section \ref{sec:model}), this experiment choice is motivated by insights in prior research \cite{nguyen2022begin}, which underscored two key points:
(1) initiating the model from pre-trained weights accentuates the necessity for an adaptive optimization on the \textit{server side} than any methods for addressing heterogeneity on the client side and (2) pre-trained models help alleviate the heterogeneity effect resulting from differences in local data, leading to better convergence.
However, as these conclusions have been validated on small-scale unimodal benchmarks, their relevance in settings with complex multimodal problems such as DocVQA, remain unclear.

\textbf{Federated Averaging (FedAvg)} \cite{mcmahan2017communication} is the standard training algorithm in federated settings. At each round of FedAvg, \text{SERVEROPT} updates its global model by averaging local updates from participating clients: $\theta_{t+1}\leftarrow \theta_{t}-\Delta^{t}$. 

\textbf{Server Momentum (FedAvgM)} \cite{hsu2019measuring} 
Incorporating momentum to Stochastic Gradient Descent, which involves accumulating the gradient history, has shown to effectively stabilize training deep neural networks. This approach appears particularly relevant in FL, when data distributions across clients are significantly different, leading to high variances in gradients. In this case, instead of directly using $\Delta^{t}$, the global model is updated by \text{SERVEROPT} as: $\theta_{t+1}\leftarrow \theta_{t}-m_t$, where $m_t = \beta m_{t-1}+(1-\beta)\Delta^{t}$ indicates the server momentum after local training at round $t$.

\textbf{Server Adaptive Optimizer (FedAdam)} \cite{reddi2020adaptive} The square of the gradient is integrated to effectively control its variance in each update, thus accelerating convergence. The update rule of \text{SERVEROPT} is then specified as follows: $\theta_{t+1}\leftarrow \theta_{t}-\eta_{s}\frac{m_t}{\sqrt{v_t}+\epsilon}$ where $v_t=\beta_{2}v_{t-1}+(1-\beta_{2})(\Delta^{t})^2$ is the second moment of the gradient \cite{kingma2014adam}.

\subsection{Federated Self-Pretraining (FSP)}
\textbf{Self-Pretraining} refers to pretraining a model on the unlabeled training data for a given task before finetuning using the labels and the task-specific loss. This section explores the applicability of this method in a multimodal-FL setting, with DocVQA as a specific usecase. Our approach to FSP is straightforward, which continues pretraining a PLM (Section \ref{sec:model}) on unlabeled documents in DocVQA datasets, in order to obtain a domain-adapted initialization for the subsequent training. Specifically, this pretraining phase is also conducted in a federated manner, with each client performing self-supervised training on its \textit{private} documents. Our assumption is that: while documents for DocVQA are heterogeneous across clients, the overall capacity to understand documents is homogeneous. Documents from many DocVQA datasets typically share attributes such as relative layout structure, reading order, etc. that are essential for DU, hence the proposed pretraining is necessary.

The choice of FSP is motivated by two reasons: (1) under such privacy constraints where centralization of documents is prohibited, local self-supervised pretraining with federation naturally fits to this setting (2) while empirically showing competitive performances \cite{el2021large,he2022masked}, it is much less expensive compared to large-scale methods \cite{tang2023unifying,borchmann2021due} that require heavy training on large corpora \cite{10.1145/1148170.1148307}.
\noindent
\textbf{Pretraining Objectives}. We describe our self-supervised tasks used in FSP, which is a set of unified generative tasks inspired from UDOP \cite{tang2023unifying}. As we use a PLM which was solely trained on text data, this local pre-training aims to effectively learn the alignment between the semantic and layout information from documents \textit{with no QA annotations}. 
Essentially, all of the tasks are based on the denoising objective proposed in T5 \cite{raffel2020exploring}, which randomly masks the multimodal input and forces the model to predict the masked information.

More formally, suppose we have one training example as input: a document image $I$, a list of text tokens $T = \{t_{1},...,t_{j}\}$ and their corresponding bounding boxes $B = \{b_{1},...,b_{j}\}$, where $b_{j} = (x^{0}_{j},y^{0}_{j},x^{1}_{j},y^{1}_{j})$ indicates the 2-D spatial information. Let $M$ be the set of indices $m$ sampled from $\{0,1,...,j\}$ with probability $p_m$ and $l_M$ be the size of $M$ which takes the maximum value $L_M$. 
Like UDOP, we define a set of sentinel tokens including: task-specific tokens as \textless task\_$l$\textgreater  where $l\in \{0,..,l_{M}\}$ to indicate the masked positions for each task; and coordinate tokens \textless loc\_$l$\textgreater  where $l\in \{0,1,..,|V|\}$ and $|V|$ is the size of a discretization $f_d$ applied on the normalized bounding box coordinates. The bounding boxes for sentinel tokens are all set to 0.
Unlike UDOP, we perform masking at \textit{token-level} only, because DocVQA with complex layouts like tables, databases, etc. requires reasoning on positional structures which are defined over fine-grained elements. We briefly summarize the structure for each objective as follows: 

\textbf{Text Modeling (TM)} requires the model to predict the masked text at given locations, thus understands the semantics of the document content based on relative positions. For each $m\in M$, we replace $t_m$ from the input with $\textless \text{text}\_l\textgreater\textless \text{loc}\_f_d(x^{0}_{m})\textgreater \textless\text{loc}\_f_d(y^{0}_{m})\textgreater\textless \text{loc}\_f_d(x^{1}_{m})\textgreater\textless \text{loc}\_f_d(y^{1}_{m}))\textgreater$ and concatenate $\textless text\_l\textgreater t_m$ to the output sequence. We use $p_m=0.5$ and set $L_M$ as 100 for this task.

\textbf{Layout Modeling (LM)} tasks the model to predict the location of given tokens, enabling it to understand the document's layout. Specifically, for each $m\in M$, we surround $t_m$ with layout tokens as $\textless\text{layout}\_l\textgreater t_m\textless\slash\text{layout}\_l\textgreater$ in the input and append $\textless\text{layout}\_l\textgreater\textless\text{loc}\_f_d(x^{0}_{m})\textgreater \textless\text{loc}\_f_d(y^{0}_{m})\textgreater\textless \text{loc}\_f_d(x^{1}_{m})\textgreater\textless \text{loc}\_f_d(y^{1}_{m})\textgreater$ to the output sequence. In this task, we use $p_m=0.75$ and set $L_M=100$ for this task.

\textbf{Text-Layout Modeling (TLM)} encourages the model to fully utilize the layout information by masking both the text $t_m$ and $b_m$ and let the model predict given the available context. In particular, we replace each $t_m$ with a single token $\textless\text{text\_layout}\_l\textgreater$ and train the model to generate all of associated information $\textless\text{text\_layout}\_l\textgreater t_m \textless \text{loc}\_f_d(x^{0}_{m})\textgreater \textless\text{loc}\_f_d(y^{0}_{m})\textgreater\textless \text{loc}\_f_d(x^{1}_{m})\textgreater\textless \text{loc}\_f_d(y^{1}_{m})\textgreater$. We set $p_m=0.15$ and $L_M=100$ for this task.
\section{Experimental Setup}
\subsection{Dataset Selection}
Real-world FL involves training the model on client's data which are generated from different distributions. A real-world example might involve considering each client as the datacenter of a different business, where documents with specific content, layout, query types etc are treated.
To this end, we consider 
making use of existing datasets of different QA types, which indeed share the same document-based reasoning structure. 
We choose one dataset as representative for each type of QA task to strengthen the data heterogeneity in our FL setting. We prioritise datasets that can be recast as end-to-end document-based QA. The datasets used are the following:

\noindent
\textbf{\docvqa} \cite{mathew2021docvqa}. A large-scale dataset comprising mostly extractive Visual QAs on single-page real-world administrative document images that include tables, forms, figures etc. Several data curation and quality control steps ensure a high quality and diversity across documents and questions.

\noindent
\textbf{WikiTableQuestions(\wtq)} \cite{pasupat-liang-2015-compositional}. This Table QA dataset comprises a collection of logical questions over HTML tables sourced from Wikipedia. The complexity of the questions demands semi-structured document understanding, compositional reasoning on a series of the table's entries, including a variety of operations such as comparison, aggregation, etc.

\noindent
\textbf{\tabfact} \cite{Chen2020TabFact:}. A big Natural Language Inference (NLI) dataset with high-quality human annotations is designed for the task of fact verification given evidence in the form of Wikipedia tables, graphs, etc. Despite its binary classification nature with \textit{yes/no} answers, verifying whether a textual hypothesis is entailed or refuted based on the evidence, as an query, requires complex linguistic and symbolic reasoning capabilities.

Table \ref{tab:data_stats} provides the properties of selected datasets in terms of their size, task and evaluation metric. For more details of these datasets, we refer the readers to the original papers. We make use of the official split provided for each dataset.
We follow the task processing, reformulation and data split as outlined in the DUE benchmark \cite{borchmann2021due} to make all the tasks aligned with the DU paradigm. Particularly, we unify all training data into the triplet (document image, question, answer) format. In addition, annotations are partly inherited from DUE, including: document image, question/answer, OCR tokens, while the normalized bounding boxes are generated the same way as in \cite{tito2023privacy}.
\input{tables/data_stats}

\subsection{Data Partition}
\label{sec:data_partition}
In FeDocVQA, we assume 
that the training examples of each client are disjoint subsets of documents from \textit{one single dataset}. To synthesize a population of non-identical $K$ clients, for each dataset $D_{doc}\in \{D_{\wtq}; D_{\docvqa}; D_{\tabfact}\}$, we evenly distribute documents into $k_{doc}\in \{k_{\wtq}; k_{\docvqa}; k_{\tabfact}\}$ clients by random sampling without replacement, such that $k_{\wtq} + k_{\docvqa} + k_{\tabfact} = K$. This strategy creates a balanced population among clients which receive the data from the same source, while ensures the data in every client unique. We consider 3 FL scenarios, which correspond to having $K \in \{3; 10; 30\}$ clients. $K=3$ signifies each client holds one entire original dataset, which is highly representative of real-world FL scenarios, whereas $K=30$ scales the problem with a high number of participating clients that each has access to a much smaller number of samples. 
Table \ref{tab:data_partition} illustrates data populations resulted from our partitioning scheme and its statistics.
\input{tables/data_partition}

\subsection{Model}
\label{sec:model}
\begin{figure}[t]
    \centering
    \includegraphics[width=0.9\linewidth]{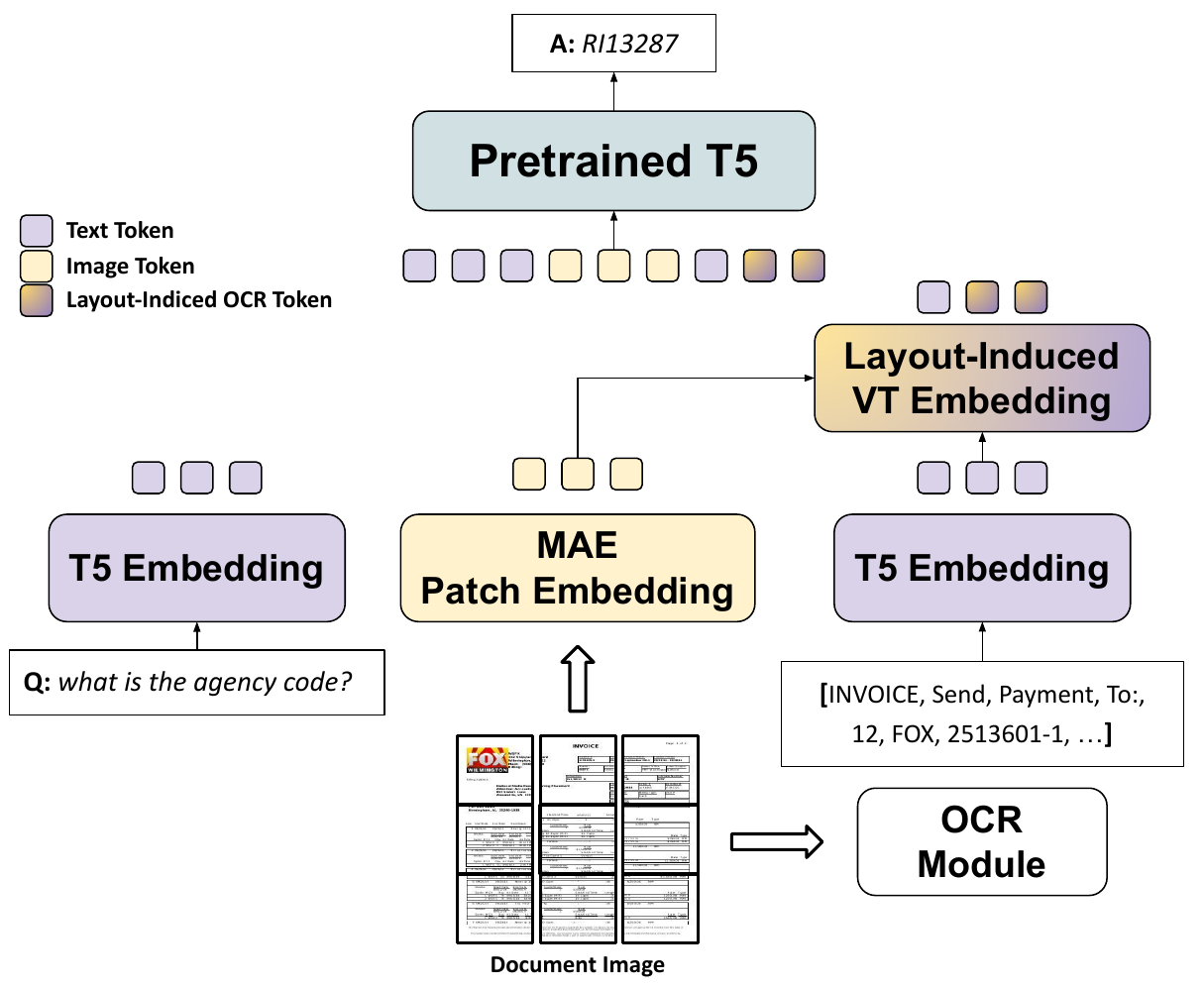}
    \caption{\textbf{The overall architecture} of the proposed model in our experiments.}
    \label{fig:model}
    \vspace{-0.75cm}
\end{figure}

In our experiments, we employ a generative model (Figure \ref{fig:model}) which unifies all QA tasks to multimodal-input-to-text generation. Specifically, we use T5 \cite{raffel2020exploring} as the backbone, initialized with \textit{t5-base} pre-trained weights. To extend the PLM to multimodal input, we augment the input with visual features from document images by reusing the patch embedding provided from \textit{mae\_vit\_base\_patch16} MAE \cite{he2022masked} checkpoints. Concretely, given the document image $I$ with the size of $H\times W\times C$ as the height, width and number of channels respectively, we break it into $n$ 2D patches of $(P\times P)$ resolution while keeping the channel dimension, such that $n=H\times W\slash P^2$. We employ MAE's patch embedding layer to transform each flattened patch into a D-dimensional space. OCR and question tokens T and Q are encoded using T5 embedding layer with the same dimension. The bounding box is normalized to a range of $[0,1]$ and embedded by a linear layer before being added to its respective text features. Following UDOP, we incorporate the patch features into the text features for any text detected within that patch region, subsequently excluding it from the set of image features. This operation is referred to as Layout-Induced Vision-Text Embedding. As a result, the T5-Encoder takes a sequence of image embeddings, OCR embeddings enriched with 2D visual/spatial information and question embeddings as input, in order to generate answers. 
We utilize the \textit{t5-base} pre-trained version from HuggingFace \cite{wolf-etal-2020-transformers} for the model and tokenizer, with sentinel tokens included.
\section{Results}
\subsection{Baselines}
\label{sec:baseline}
Given the dataset preparation outlined in Section \ref{sec:data_partition}, we evaluate the performance of the FedAvg algorithm with few modifications, which serves as the baseline for FeDocVQA.

We employ the Adam optimizer \cite{kingma2014adam} as CLIENTOPT (instead of SGD in vanilla FedAvg) with fixed parameters across all clients: local learning rate $\eta_l=0.0005$ and weight decay of 0.01. We reset the optimizer at the beginning of each round and opt not to use learning rate scheduler, since it requires to keep track of local states across rounds. We fix the momentum coefficient $\beta_1 = 0.9$ and a second moment coefficient $\beta_2 = 0.999$. FedAvg is run with client batch size $B = 16$, local epochs $E = 1$ and sampling probability
$C \in \{0.35; 0.7; 1\}$ (corresponding to $0.35\cdot K$, $0.7\cdot K$ and $K$ clients participating in each round, respectively) for a total of $T=10$ rounds. 
Performance metrics are reported as a two-step average: first, we compute the average score of test examples for each dataset, and then average per-dataset scores to obtain the final score.
\input{tables/baseline}

Table \ref{tab:baseline} summarizes FeDocVQA performance of FedAvg as a function of number of total clients $K$ and sampling probability $C$. Plots of validation performances (loss/metric) for the case $K=3$ are in Figure \ref{fig:baseline_val}.
We also present results for $(C=1, K=1)$, which corresponds to the centralized training, where the model is finetuned using data from all datasets collectively. Compared to the T5-large variant \cite{borchmann2021due}, simply finetuning our model already achieves comparable performance with fewer parameters (220M vs 770M).

We first analyze the result with different values of $K$. We observe a significantly lower performance when $K=3$ and $C=0.35$, corresponding to fine-tuning the model on one dataset after another, leading to catastrophic forgetting. It can also be seen from Figure \ref{fig:baseline_val} and Figure \ref{fig:baseline_val_breakdown} that the validation loss/per-dataset metric fluctuates considerably at each round.
Interestingly, increasing in the number of clients $K$ from 3 to 10 yields substantial improvements, but further increasing from 10 to 30 results in a decline in performance. For instance, considering $C=0.35$, increasing $K$ from 3 to 10 highlights the advantages of FL, as the server aggregates training updates from multiple clients, therefore mitigating heterogeneity in each communication round. However, the performance drop observed when increasing $K$ from 10 to 30 is attributed to significantly less local computation and more frequent model synchronization. Specifically, recall from Table \ref{tab:data_partition}, when $K=10$, each client holds 2x-4x more data compared to when $K=30$, allowing for more gradient steps during local training and less frequent weight synchronization. In contrast, $K=30$ entails a more widely spread population with less training data per client. This observation suggests that \textit{having more local updates per round can speedup convergence and result in better performance}, aligning with prior research findings \cite{mcmahan2017communication}. Furthermore, it is worth noting that the partition is synthetically created, thus the final performance is sensitive on the client-to-dataset ratio, which is computed based on the size of each dataset.
\input{figures/baseline}

For most cases of $K$, we observe that significant improvements occur around high values of $C$, particularly with $K=3$, demonstrating that incorporating more clients per communication round has a notable impact. This result confirms our assumption that there is \textit{an inherent structure driving document understanding capabilities across various DocVQA datasets}, despite their disparate domains.
In the literature, there has not been a definitive conclusion regarding the effect of $C$ when it approaches full-participation ($C=1$). For instance, \cite{mcmahan2017communication,hsu2019measuring} reported results with only small values of $C\leq0.5$ and observed marginal improvements between $0.2$ to $0.5$. We speculate that prior studies, which mainly focused on randomly initialized models and scenarios with limited multi-client parallelism, may not generalize well to complex tasks like DocVQA, where pretrained models are extensively utilized with sufficient computation.


\input{tables/ablate_fsp}
\subsection{Ablation Study}
In this section, we select several factors that appear to be important in our proposed setup to study their effects and provide empirical results when tuning these hyperparameters.
Unless otherwise specified, we employ the FedAvg baseline and fix $K = 3$ and $C = 0.35$ in all of our experiments.

\noindent
\textbf{Effect of FSP.} The purpose of FSP is to provide a warm start to the PLM, allowing it to better adapt to document data. In Table \ref{tab:ablate_fsp_ssl}, we compare the performance of FedAvg with various pretraining objectives. As expected, each pretraining objective improves the results compared to the baseline. When all objectives are combined, FSP achieves a great improvement in FeDocVQA, outperforming the baseline by approximately 2.6 points while solely utilizing private data, which is considerably smaller in size compared to common practices. We believe \textit{our FSP strategy can effectively scale up} with more documents while maintaining a high level of client data privacy. We utilize all pretraining tasks in our subsequent experiments. 

\noindent
\textbf{Effect of Client fraction $C$.} We also study the effect of the $C$ for each round during pretraining/finetuning, denoted as $C_{\text{pt}}$ and $C_{\text{ft}}$ respectively. Table \ref{tab:ablate_fsp_c} gives results for each value pair of $(C_{\text{pt}}, C_{\text{ft}})$, both in $\{0.35; 0.7; 1\}$. First, we see that for each $C_{\text{ft}}$ value, there is at least one configuration $(C_{\text{pt}}, C_{\text{ft}})$ that outperforms the FedAvg baseline (compare numbers in each row). This demonstrates the consistent effectiveness of FSP across varying numbers of clients participating in each round. This result holds especially when $C_{\text{pt}}$ and $C_{\text{ft}}$ have the same value, implying that \textit{maintaining a consistent parallelism between the two training phases is beneficial}. In our subsequent experiments, we employ the same values for both $C_{\text{pt}}$ and $C_{\text{ft}}$.

\noindent
\textbf{Effect of number of communication rounds $T$.}
Similarly, we investigate the impact of $T$ for both FSP and DocVQA federated training. First, we run experiments having upto 25 rounds of communication with the FedAvg baseline. In another set of experiments, we run FSP with varying number of $T$ while fixing this value for finetuning to 10. This choice is made because FSP is less computationally intensive than DocVQA, provided that training data size for FSP is significantly smaller compared to DocVQA, as indicated in Table \ref{tab:data_stats}.
\input{tables/ablate_t_serveropt}
From Figure \ref{fig:ablate_t} (right), there is a substantial improvement while increasing the communication rounds $T$ to 10 in FSP, with further increases resulting in only marginal improvements. A similar trend is also noticed in Figure \ref{fig:ablate_t} (left) as the finetuning round $T$ increases.
We hypothesize that in the early stages of training, the drift in local updates prevents the model from learning effectively. 
This necessitates a \textit{much longer period to achieve convergence}, therefore increasing the overall cost of training. Based on these results, for our next experiments we opt to fix $T = 10$, which strikes a good balance between communication efficiency and DocVQA performance.

\noindent
\textbf{Effect of Server Optimizer.} In these experiments, we apply adaptive optimization methods as described in Section \ref{sec:fl}: FedAvgM and FedAdam. We only utilize adaptive optimization as SERVEROPT during FSP while fixing it to FedAvg during finetuning. The choice is motivated by our previous experiments where we show obtaining better pretrained weights leads to better performance for the federated training. We use momentum $\beta=0.9$ for FedAvgM, and $\beta_1=0.9, \beta_2=0.99 $ and $\epsilon$ is $1e\text{-}5$ as the momentum hyperparameters for FedAdam. We keep the server learning rate $\eta_s$ at 0.001.

Table \ref{tab:ablate_serveropt} and Figure \ref{fig:loss_fsp} show the effect of SERVEROPT with adaptive server optimizers compared to the FedAvg baseline. Overvall, FedAdam yields better performance than FedAvgM, since it has a mechanism to better adapt the learning rate, thus achieves better convergence. When we consider more clients with $K=10$, we observe that using FedAdam constantly outperforms FedAvg, suggesting that FedAdam is more effective to deal with a high level of heterogeneity. However, we suggest that performances for both strategies can be better improved with extensive hyperparameter tuning, for such complex setting like DocVQA with FL as it involves numerous hyperparameters.

\subsection{Main Results}
Table \ref{tab:main_results} presents our main results over our set of proposed methods and experiments. We select the best results, based on the hyperparameters selected for ablation study, for each configuration $(K,C)$. Overall, our FSP strategy is beneficial for DocVQA under FL with consistent improvements across different configurations. While FedAvg can still be effective for small-scale setting like $K=3$, FedAdam is considered the better design choice in a more heterogeneous system like $K=10$.
\input{tables/main_results}

\section{Conclusions}
In this work we showed how to apply a federated learning scheme to train a DocVQA model on decentralised document data, stemming from three different, heterogeneous sources. We demonstrated that FL is a viable approach for both pre-training and fine-tuning large-scale multimodal LLM models like the ones used for DocVQA, reaching results comparable to the centralised model. This can potentially enable researchers to take advantage of document collections scattered across private data silos, which can lead to better generalisation as it allows training over more heterogeneous data.

\section*{Acknowledgements}
\sloppy
This work has received support from Grants PID2020-116298GB-I00 and PLEC2021-007850 funded by MCIN/AEI/10.13039/501100011033 and the project ELSA, 101070617, financed by the European Union’s Horizon Europe research and innovation programme.

%
%
%
\bibliographystyle{splncs04}
\bibliography{main}
\end{document}

%% file: preamble.tex
\newcommand{\wtq}{WTQ\xspace}
\newcommand{\docvqa}{DocVQA\xspace}
\newcommand{\tabfact}{TabFact\xspace}

%% file: figures/teaser.tex
\begin{figure}[h!]
    \centering
    \includegraphics[width=0.7\linewidth]{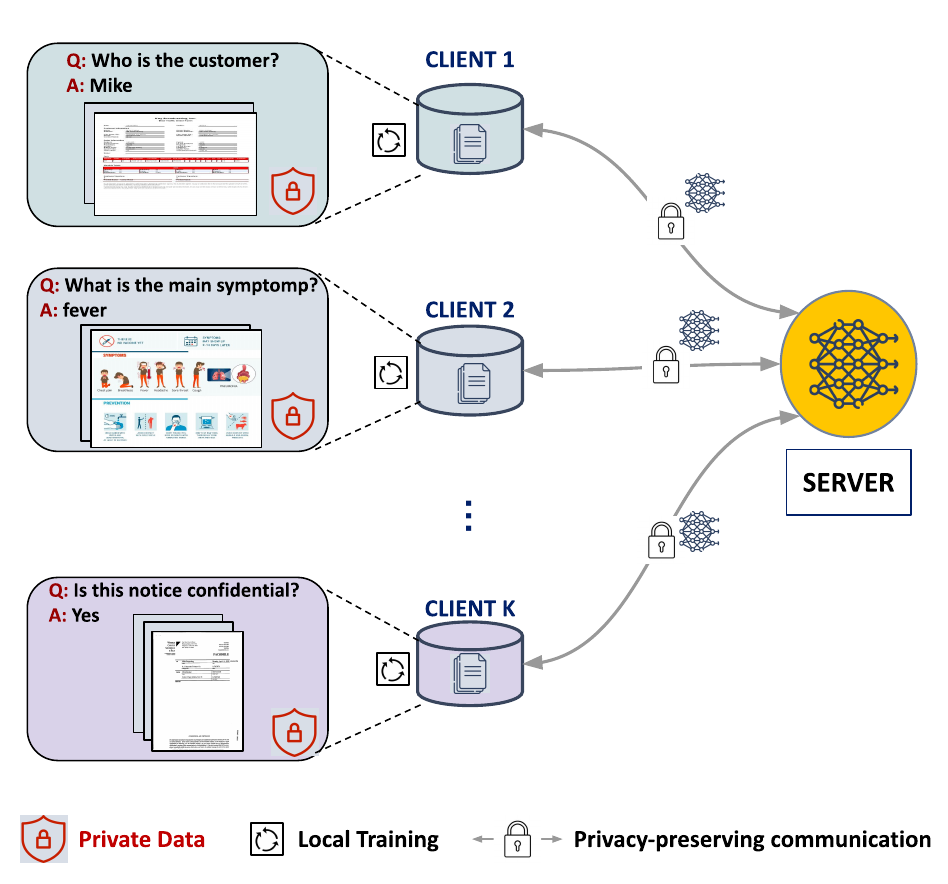}
    \caption{\textbf{The general scheme of Federated Document Visual Question Answering}. The fundamental concept involves collaborative training of a DocVQA model under the coordination of the server. The training takes place locally at each client and \textit{only model updates} are communicated between client and server, ensuring that private data is \textit{never shared}, thus preserving privacy.}
    \label{fig:fedocvqa}
    \vspace{-0.8cm}
\end{figure}

%% file: tables/data_stats.tex
\begin{table}[h]
  \centering
  \begin{tabular}{lcccccccccc}
    \toprule
    \multirow{2}{*}{Dataset} & \multirow{2}{*}{QA Type} & \multicolumn{2}{c}{train}  & \multicolumn{2}{c}{val}  & \multicolumn{2}{c}{test}  & \multicolumn{2}{c}{total} & \multirow{2}{*}{Metric}\\
    \cmidrule(r){3-10}
    & & \#D & \#Q & \#D & \#Q & \#D & \#Q & \#D & \#Q & \\
    \midrule
    \wtq & Table QA & 1.3 & 14.2 & 0.3 & 3.5 & 0.4 & 4.3 & 2.1 & 21.9 & anls\\								
    \docvqa & Visual QA  & 10.2 & 39.4 & 1.3 & 5.4 & 1.3 & 5.2 & 12.8 & 49.9 & anls\\
    \tabfact & Table NLI & 13.2 & 91.8 & 1.7 & 12.7 & 1.7 & 12.7 & 16.6& 117.2 & accuracy\\
    \bottomrule
    \tiny
  \end{tabular}
  \caption{\textbf{Summary of existing QA datasets}, following the reformulation and split provided from DUE \cite{borchmann2021due}. \#D/\#Q denotes number of documents/questions in each dataset respectively. Numbers are reported in thousands.}
  \label{tab:data_stats}
  \vspace{-1.5cm}
\end{table}

%% file: tables/data_partition.tex
\begin{table}[t]
    \centering
    \begin{tabular}{lccc}
    \toprule
    & $K=3$ & $K=10$ & $K=30$\\
    \midrule 
    \wtq & \textbf{1}(1346) & \textbf{1}(1346) & \textbf{2}(673)\\								
    \docvqa & \textbf{1}(10194) & \textbf{4}(2548) & \textbf{13}(784)\\								
    \tabfact & \textbf{1}(13163) & \textbf{5}(2632) & \textbf{15}(877) \\
    \bottomrule
    \tiny
  \end{tabular}
  \caption{\textbf{Number of participating clients in each split strategy.} In parenthesis, the average number of training documents per client. In our FL partitions, each holds data only from one DocVQA dataset.}
  \label{tab:data_partition}
  \vspace{-0.8cm}
\end{table}

%% file: tables/baseline.tex

\begin{table}[t]
\begin{minipage}{.58\linewidth}
  \centering
  \begin{tabular}[t]{lcccc}
    \toprule
    T5-large\cite{borchmann2021due} & \multicolumn{4}{c}{54.2} \\
    \midrule 
    $C=1.0$ & 53.16 & 46.54 & 45.67 & 37.64\\
    $C=0.7$ & - & 43.3 & 43.21 & 35.21\\
    $C=0.35$ & - & 35.04 & 41.66 & 32.42\\
    \midrule 
    & $K=1$ & $K=3$ & $K=10$ & $K=30$\\ 			
    \bottomrule
  \end{tabular}
  \medskip
  \caption{\textbf{FedAvg performance on FeDocVQA test set}. Each cell corresponds to one $(K,C)$ configuration. The centralized learning results of T5-large is based on DUE Benchmark \cite{borchmann2021due}. Metric is computed as two-stage average.}
  \label{tab:baseline}
\end{minipage}
\hfill
\begin{minipage}{.38\linewidth}
\centering
\begin{tabular}[t]{cccc}
    \toprule
    LM & TM & TLM & Metric\\
    \midrule
    \multicolumn{3}{c}{No FSP} & 35.04\\
    \midrule
    $\checkmark$ & $\checkmark$ &  & 35.76\\			 
    &  & $\checkmark$ & 36.74\\			
    $\checkmark$ & $\checkmark$ & $\checkmark$ & \textbf{37.65}\\
    \bottomrule
\end{tabular}
\medskip
\caption{\textbf{FedAvg performance on FeDocVQA test set with FSP} using different pretraining objectives. We run $T=10$ pretraining rounds.}
\label{tab:ablate_fsp_ssl}
\end{minipage}
\vspace{-0.8cm}
\end{table}

%% file: figures/baseline.tex
\begin{figure}[t]
\begin{minipage}[c]{0.64\linewidth}
\centering
    \includegraphics[width=0.9\linewidth]{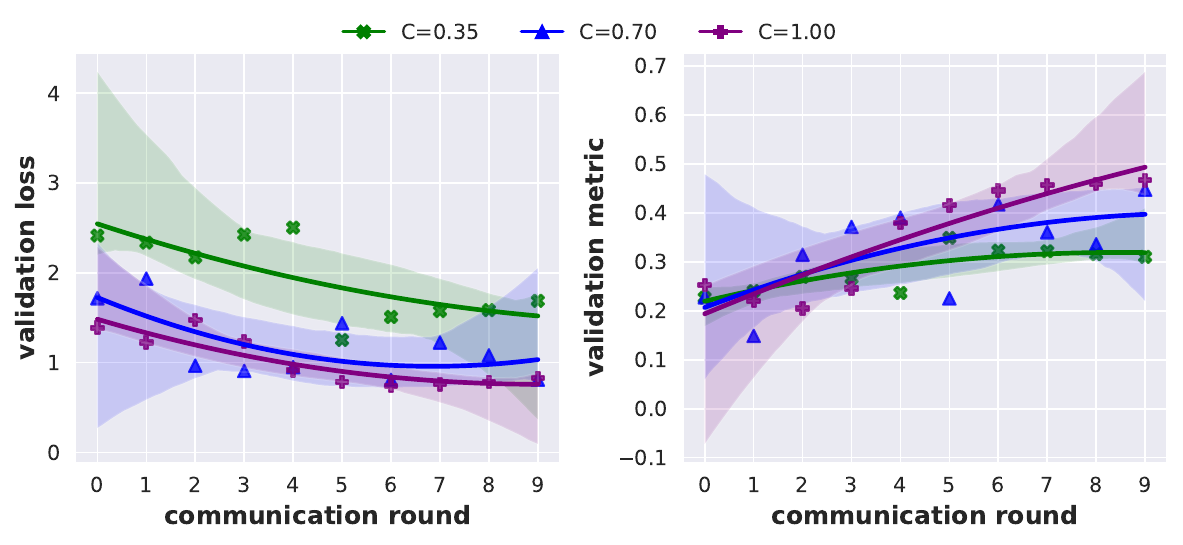}
    \caption{\textbf{FedAvg training progress with Adam optimizer as CLIENTOPT}. This plot presents the validation loss (left) and metric (right) curves for $K=3$ while varying $C$. We fit a quadratic regression model for better visualization.}
    \label{fig:baseline_val}
\end{minipage}
\hfill
\begin{minipage}[c]{0.33\linewidth}
\centering
    \includegraphics[width=\linewidth]{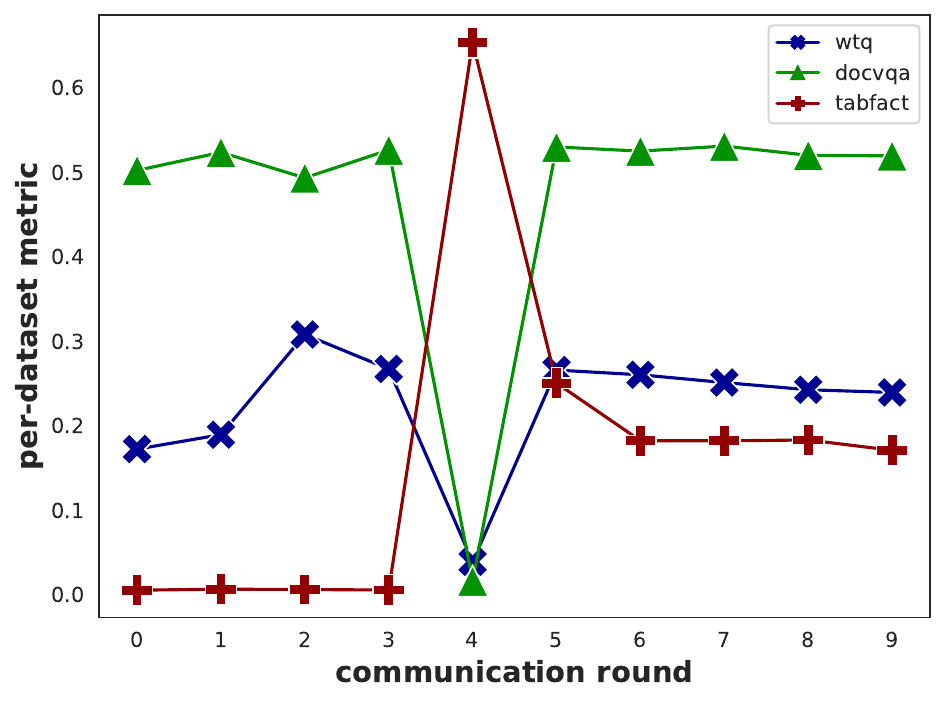}
\caption{\textbf{Breakdown of per-dataset metrics} on validation set over FedAvg training, with $K=3$ and $C=0.35$.}
\label{fig:baseline_val_breakdown}
\end{minipage}
\vspace{-0.6cm}
\end{figure}

%% file: tables/ablate_fsp.tex
\definecolor{Gray}{rgb}{0.54, 0.81, 0.94}
\begin{table}[t]
\begin{minipage}{.51\linewidth}
\centering
\begin{tabular}[t]{lccccc}
    \toprule
    & & \multicolumn{3}{c}{FSP+FedAvg} & FedAvg\\
    \midrule
    \multirow{3}{*}{$C_{\text{ft}}$}& $1.0$ & 46.65 & 47.92 & \textbf{48.66} & 46.54 \\
    & $0.7$ & 42.59 & \textbf{44.54} & 43.1 & 43.3\\
    & $0.35$ & \textbf{37.65} & 36.26 & 35.6 & 35.04\\
    \midrule 
    $C_{\text{pt}}$ & & $0.35$ & $0.7$ & $1.0$ &  \\
    \bottomrule
  \end{tabular}
\medskip
\caption{\textbf{Performance on FeDocVQA test set of FedAvg+FSP when varying Client fraction $C$}. We make use of all pretraining tasks in these experiments. $C_{\text{pt}}$/$C_{\text{ft}}$ is the clients-per-round fraction in used in pretraining and finetuning respectively.}
\label{tab:ablate_fsp_c}
\end{minipage}
\hfill
\begin{minipage}[c]{0.48\textwidth}
    \centering
    \includegraphics[width=\linewidth]{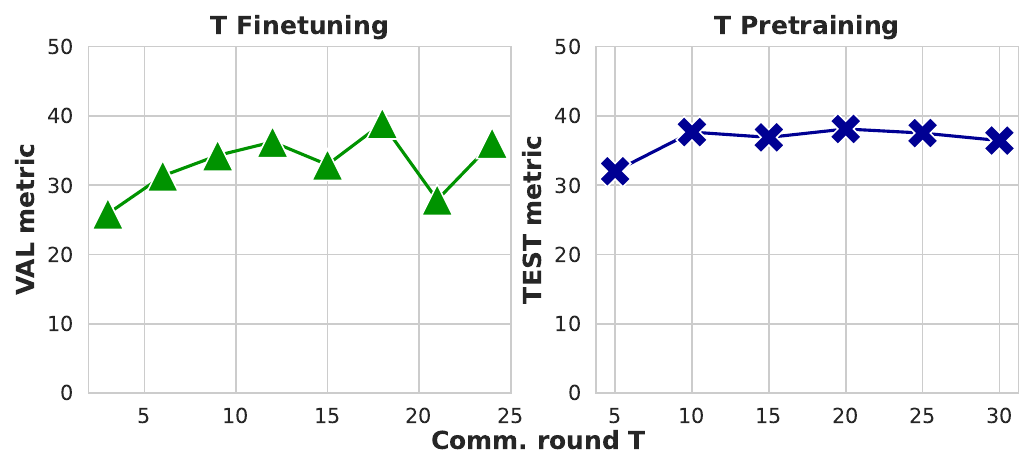}
    \captionof{figure}{\textbf{Validation/Test performance vs. communication rounds $T$}. (Left) validation metric of FedAvg baseline while varying T in \textit{finetuning}, (Right) test metric of FSP+FedAvg while varying T in \textit{pretraining}.}
    \label{fig:ablate_t}
\end{minipage}
\vspace{-0.8cm}
\end{table}

%% file: tables/ablate_t_serveropt.tex
\definecolor{Gray}{rgb}{0.54, 0.81, 0.94}
\begin{minipage}{\textwidth}
 \begin{minipage}[b]{0.49\textwidth}
    \centering
    \includegraphics[width=0.92\linewidth]{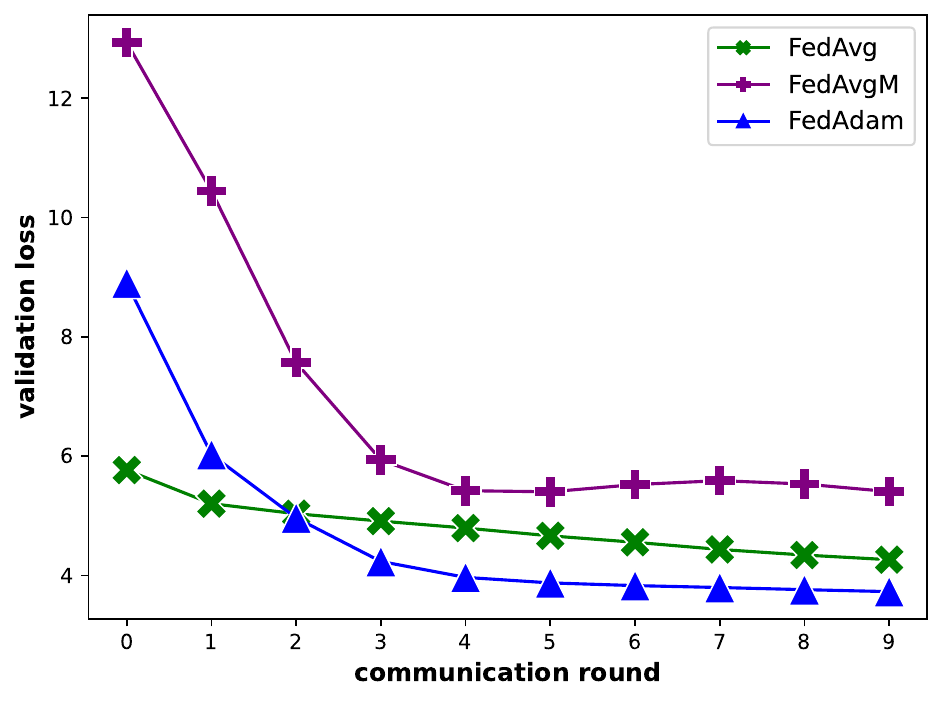}
    \captionof{figure}{\textbf{Validation FSP loss} with different SERVEROPT algorithms, in case of $(K,C)=(10,0.7)$.}
    \label{fig:loss_fsp}
  \end{minipage}
  \hfill
  \begin{minipage}[b]{0.49\textwidth}
    \centering
    \begin{tabular}{cccc}
    \toprule
    \multirow{2}{*}{$(K, C)$} & \multicolumn{2}{c}{FSP+FedAvg} & \multirow{2}{*}{FedAvg}\\ 
     \cmidrule{2-3}
     & FedAvgM & FedAdam & \\ 
     \midrule
    $(3,0.35)$ & 35.72 & \textbf{36.05} & 35.04\\ 
    $(3,0.7)$ & 37.25 & 39.82 & \textbf{43.33}\\
    $(10,0.35)$ & 40 & \textbf{43.93} & 41.66\\
    $(10,0.7)$ & 42.7 & \textbf{45.2} & 43.21\\
    \bottomrule
    \end{tabular}
      \captionof{table}{\textbf{FedAvgM and FedAdam Test set performance on different configurations of $(K,C)$.} The rightmost column shows the baseline's performance.}
      \label{tab:ablate_serveropt}
    \end{minipage}
\end{minipage}

%% file: tables/main_results.tex
\begin{table}[t]
  \centering
  \begin{tabular}{lccc}
    \toprule
    & FSP(FedAvg) & FSP(FedAdam) & FSP(FedAdam)\\
    & +FedAvg & +FedAvg & +FedAvg\\
    \midrule
    $C=0.7$  & \textbf{44.54}(+2.12) & \textbf{45.2}(+1.99) & \textbf{38.55}(+3.34)\\
    $C=0.35$ & \textbf{37.65}(+2.61) & \textbf{43.93}(+2.27) & \textbf{33.06}(+0.64)\\
    \midrule 
    & $K=3$ & $K=10$ & $K=30$\\		
    \bottomrule
    \tiny
  \end{tabular}
  \caption{\textbf{Main results of our proposed methods on FeDocVQA test set}. In parenthesis, we provide the performance gain compared to the baseline. Each cell corresponds to one $(K,C)$ configuration.}
  \label{tab:main_results}
  \vspace{-0.8cm}
\end{table}

%% file: main.bbl
\begin{thebibliography}{10}
\providecommand{\url}[1]{\texttt{#1}}
\providecommand{\urlprefix}{URL }
\providecommand{\doi}[1]{https://doi.org/#1}

\bibitem{borchmann2021due}
Borchmann, {\L}., Pietruszka, M., Stanislawek, T., Jurkiewicz, D., Turski, M., Szyndler, K., Grali{\'n}ski, F.: {DUE}: End-to-end document understanding benchmark. In: Thirty-fifth Conference on Neural Information Processing Systems Datasets and Benchmarks Track (Round 2) (2021), \url{https://openreview.net/forum?id=rNs2FvJGDK}

\bibitem{che2023multimodal}
Che, L., Wang, J., Zhou, Y., Ma, F.: Multimodal federated learning: A survey. Sensors  \textbf{23}(15), ~6986 (2023)

\bibitem{Chen2020TabFact:}
Chen, W., Wang, H., Chen, J., Zhang, Y., Wang, H., Li, S., Zhou, X., Wang, W.Y.: Tabfact: A large-scale dataset for table-based fact verification. In: International Conference on Learning Representations (2020), \url{https://openreview.net/forum?id=rkeJRhNYDH}

\bibitem{dayan2021federated}
Dayan, I., Roth, H.R., Zhong, A., Harouni, A., Gentili, A., Abidin, A.Z., Liu, A., Costa, A.B., Wood, B.J., Tsai, C.S., et~al.: Federated learning for predicting clinical outcomes in patients with covid-19. Nature medicine  \textbf{27}(10),  1735--1743 (2021)

\bibitem{5206848}
Deng, J., Dong, W., Socher, R., Li, L.J., Li, K., Fei-Fei, L.: Imagenet: A large-scale hierarchical image database. In: 2009 IEEE Conference on Computer Vision and Pattern Recognition. pp. 248--255 (2009). \doi{10.1109/CVPR.2009.5206848}

\bibitem{dosovitskiy2020image}
Dosovitskiy, A., Beyer, L., Kolesnikov, A., Weissenborn, D., Zhai, X., Unterthiner, T., Dehghani, M., Minderer, M., Heigold, G., Gelly, S., et~al.: An image is worth 16x16 words: Transformers for image recognition at scale. arXiv preprint arXiv:2010.11929  (2020)

\bibitem{el2021large}
El-Nouby, A., Izacard, G., Touvron, H., Laptev, I., Jegou, H., Grave, E.: Are large-scale datasets necessary for self-supervised pre-training? arXiv preprint arXiv:2112.10740  (2021)

\bibitem{ghazi2019scalable}
Ghazi, B., Pagh, R., Velingker, A.: Scalable and differentially private distributed aggregation in the shuffled model. arXiv preprint arXiv:1906.08320  (2019)

\bibitem{pmlr-v130-girgis21a}
Girgis, A., Data, D., Diggavi, S., Kairouz, P., Theertha~Suresh, A.: Shuffled model of differential privacy in federated learning. In: Banerjee, A., Fukumizu, K. (eds.) Proceedings of The 24th International Conference on Artificial Intelligence and Statistics. Proceedings of Machine Learning Research, vol.~130, pp. 2521--2529. PMLR (13--15 Apr 2021), \url{https://proceedings.mlr.press/v130/girgis21a.html}

\bibitem{gururangan2020don}
Gururangan, S., Marasovi{\'c}, A., Swayamdipta, S., Lo, K., Beltagy, I., Downey, D., Smith, N.A.: Don't stop pretraining: Adapt language models to domains and tasks. arXiv preprint arXiv:2004.10964  (2020)

\bibitem{hamer2020fedboost}
Hamer, J., Mohri, M., Suresh, A.T.: Fedboost: A communication-efficient algorithm for federated learning. In: International Conference on Machine Learning. pp. 3973--3983. PMLR (2020)

\bibitem{he2022masked}
He, K., Chen, X., Xie, S., Li, Y., Doll{\'a}r, P., Girshick, R.: Masked autoencoders are scalable vision learners. In: Proceedings of the IEEE/CVF conference on computer vision and pattern recognition. pp. 16000--16009 (2022)

\bibitem{hsu2019measuring}
Hsu, T.M.H., Qi, H., Brown, M.: Measuring the effects of non-identical data distribution for federated visual classification. arXiv preprint arXiv:1909.06335  (2019)

\bibitem{huang2022layoutlmv3}
Huang, Y., Lv, T., Cui, L., Lu, Y., Wei, F.: Layoutlmv3: Pre-training for document ai with unified text and image masking. In: Proceedings of the 30th ACM International Conference on Multimedia. pp. 4083--4091 (2022)

\bibitem{jiang2020federated}
Jiang, J.C., Kantarci, B., Oktug, S., Soyata, T.: Federated learning in smart city sensing: Challenges and opportunities. Sensors  \textbf{20}(21), ~6230 (2020)

\bibitem{kairouz2021advances}
Kairouz, P., McMahan, H.B., Avent, B., Bellet, A., Bennis, M., Bhagoji, A.N., Bonawitz, K., Charles, Z., Cormode, G., Cummings, R., et~al.: Advances and open problems in federated learning. Foundations and Trends{\textregistered} in Machine Learning  \textbf{14}(1--2),  1--210 (2021)

\bibitem{karimireddy2020scaffold}
Karimireddy, S.P., Kale, S., Mohri, M., Reddi, S., Stich, S., Suresh, A.T.: Scaffold: Stochastic controlled averaging for federated learning. In: International conference on machine learning. pp. 5132--5143. PMLR (2020)

\bibitem{Kim22Donut}
Kim, G., Hong, T., Yim, M., Nam, J., Park, J., Yim, J., Hwang, W., Yun, S., Han, D., Park, S.: Ocr-free document understanding transformer. In: Avidan, S., Brostow, G., Ciss{\'e}, M., Farinella, G.M., Hassner, T. (eds.) Computer Vision -- ECCV 2022. pp. 498--517. Springer Nature Switzerland, Cham (2022)

\bibitem{kingma2014adam}
Kingma, D.P., Ba, J.: Adam: A method for stochastic optimization. arXiv preprint arXiv:1412.6980  (2014)

\bibitem{krishna2022downstream}
Krishna, K., Garg, S., Bigham, J.P., Lipton, Z.C.: Downstream datasets make surprisingly good pretraining corpora. arXiv preprint arXiv:2209.14389  (2022)

\bibitem{Lee23Pix2Struct}
Lee, K., Joshi, M., Turc, I., Hu, H., Liu, F., Eisenschlos, J., Khandelwal, U., Shaw, P., Chang, M.W., Toutanova, K.: Pix2struct: screenshot parsing as pretraining for visual language understanding. In: Proceedings of the 40th International Conference on Machine Learning. ICML'23, JMLR.org (2023)

\bibitem{10.1145/1148170.1148307}
Lewis, D., Agam, G., Argamon, S., Frieder, O., Grossman, D., Heard, J.: Building a test collection for complex document information processing. In: Proceedings of the 29th Annual International ACM SIGIR Conference on Research and Development in Information Retrieval. p. 665–666. SIGIR '06, Association for Computing Machinery, New York, NY, USA (2006). \doi{10.1145/1148170.1148307}, \url{https://doi.org/10.1145/1148170.1148307}

\bibitem{li2021model}
Li, Q., He, B., Song, D.: Model-contrastive federated learning. In: Proceedings of the IEEE/CVF conference on computer vision and pattern recognition. pp. 10713--10722 (2021)

\bibitem{li2020federated}
Li, T., Sahu, A.K., Zaheer, M., Sanjabi, M., Talwalkar, A., Smith, V.: Federated optimization in heterogeneous networks. Proceedings of Machine learning and systems  \textbf{2},  429--450 (2020)

\bibitem{lin2020fedrec}
Lin, G., Liang, F., Pan, W., Ming, Z.: Fedrec: Federated recommendation with explicit feedback. IEEE Intelligent Systems  \textbf{36}(5),  21--30 (2020)

\bibitem{mathew2022infographicvqa}
Mathew, M., Bagal, V., Tito, R., Karatzas, D., Valveny, E., Jawahar, C.: Infographicvqa. In: Proceedings of the IEEE/CVF Winter Conference on Applications of Computer Vision. pp. 1697--1706 (2022)

\bibitem{mathew2021docvqa}
Mathew, M., Karatzas, D., Jawahar, C.: Docvqa: A dataset for vqa on document images. In: Proceedings of the IEEE/CVF winter conference on applications of computer vision. pp. 2200--2209 (2021)

\bibitem{mcmahan2017communication}
McMahan, B., Moore, E., Ramage, D., Hampson, S., y~Arcas, B.A.: Communication-efficient learning of deep networks from decentralized data. In: Artificial intelligence and statistics. pp. 1273--1282. PMLR (2017)

\bibitem{nguyen2022begin}
Nguyen, J., Malik, K., Sanjabi, M., Rabbat, M.: Where to begin? exploring the impact of pre-training and initialization in federated learning. arXiv preprint arXiv:2206.15387  (2022)

\bibitem{pasupat-liang-2015-compositional}
Pasupat, P., Liang, P.: Compositional semantic parsing on semi-structured tables. In: Zong, C., Strube, M. (eds.) Proceedings of the 53rd Annual Meeting of the Association for Computational Linguistics and the 7th International Joint Conference on Natural Language Processing (Volume 1: Long Papers). pp. 1470--1480. Association for Computational Linguistics, Beijing, China (Jul 2015). \doi{10.3115/v1/P15-1142}, \url{https://aclanthology.org/P15-1142}

\bibitem{powalski2021going}
Powalski, R., Borchmann, {\L}., Jurkiewicz, D., Dwojak, T., Pietruszka, M., Palka, G.: Going full-tilt boogie on document understanding with text-image-layout transformer. In: Document Analysis and Recognition--ICDAR 2021: 16th International Conference, Lausanne, Switzerland, September 5--10, 2021, Proceedings, Part II 16. pp. 732--747. Springer (2021)

\bibitem{raffel2020exploring}
Raffel, C., Shazeer, N., Roberts, A., Lee, K., Narang, S., Matena, M., Zhou, Y., Li, W., Liu, P.J.: Exploring the limits of transfer learning with a unified text-to-text transformer. The Journal of Machine Learning Research  \textbf{21}(1),  5485--5551 (2020)

\bibitem{reddi2020adaptive}
Reddi, S., Charles, Z., Zaheer, M., Garrett, Z., Rush, K., Kone{\v{c}}n{\`y}, J., Kumar, S., McMahan, H.B.: Adaptive federated optimization. arXiv preprint arXiv:2003.00295  (2020)

\bibitem{sattler2019robust}
Sattler, F., Wiedemann, S., M{\"u}ller, K.R., Samek, W.: Robust and communication-efficient federated learning from non-iid data. IEEE transactions on neural networks and learning systems  \textbf{31}(9),  3400--3413 (2019)

\bibitem{tang2019doublesqueeze}
Tang, H., Yu, C., Lian, X., Zhang, T., Liu, J.: Doublesqueeze: Parallel stochastic gradient descent with double-pass error-compensated compression. In: International Conference on Machine Learning. pp. 6155--6165. PMLR (2019)

\bibitem{tang2023unifying}
Tang, Z., Yang, Z., Wang, G., Fang, Y., Liu, Y., Zhu, C., Zeng, M., Zhang, C., Bansal, M.: Unifying vision, text, and layout for universal document processing. In: Proceedings of the IEEE/CVF Conference on Computer Vision and Pattern Recognition. pp. 19254--19264 (2023)

\bibitem{tito2023hierarchical}
Tito, R., Karatzas, D., Valveny, E.: Hierarchical multimodal transformers for multipage docvqa. Pattern Recognition  \textbf{144},  109834 (2023)

\bibitem{tito2021icdar}
Tito, R., Mathew, M., Jawahar, C., Valveny, E., Karatzas, D.: Icdar 2021 competition on document visual question answering. In: International Conference on Document Analysis and Recognition. pp. 635--649. Springer (2021)

\bibitem{tito2023privacy}
Tito, R., Nguyen, K., Tobaben, M., Kerkouche, R., Souibgui, M.A., Jung, K., Kang, L., Valveny, E., Honkela, A., Fritz, M., Karatzas, D.: Privacy-aware document visual question answering. arXiv preprint arXiv:2312.10108  (2023)

\bibitem{van2023document}
Van~Landeghem, J., Tito, R., Borchmann, {\L}., Pietruszka, M., Joziak, P., Powalski, R., Jurkiewicz, D., Coustaty, M., Anckaert, B., Valveny, E., et~al.: Document understanding dataset and evaluation (dude). In: Proceedings of the IEEE/CVF International Conference on Computer Vision. pp. 19528--19540 (2023)

\bibitem{vogels2019powersgd}
Vogels, T., Karimireddy, S.P., Jaggi, M.: Powersgd: Practical low-rank gradient compression for distributed optimization. Advances in Neural Information Processing Systems  \textbf{32} (2019)

\bibitem{wolf-etal-2020-transformers}
Wolf, T., Debut, L., Sanh, V., Chaumond, J., Delangue, C., Moi, A., Cistac, P., Rault, T., Louf, R., Funtowicz, M., Davison, J., Shleifer, S., von Platen, P., Ma, C., Jernite, Y., Plu, J., Xu, C., Le~Scao, T., Gugger, S., Drame, M., Lhoest, Q., Rush, A.: Transformers: State-of-the-art natural language processing. In: Liu, Q., Schlangen, D. (eds.) Proceedings of the 2020 Conference on Empirical Methods in Natural Language Processing: System Demonstrations. pp. 38--45. Association for Computational Linguistics, Online (Oct 2020). \doi{10.18653/v1/2020.emnlp-demos.6}, \url{https://aclanthology.org/2020.emnlp-demos.6}

\bibitem{wu2022insights}
Wu, Y., Li, F., Liang, P.S.: Insights into pre-training via simpler synthetic tasks. Advances in Neural Information Processing Systems  \textbf{35},  21844--21857 (2022)

\bibitem{yang2019federated}
Yang, Q., Liu, Y., Chen, T., Tong, Y.: Federated machine learning: Concept and applications. ACM Transactions on Intelligent Systems and Technology (TIST)  \textbf{10}(2),  1--19 (2019)

\end{thebibliography}
